\documentclass[10pt,twocolumn,letterpaper]{article}

\usepackage{cvpr}
\usepackage{times}
\usepackage{epsfig}
\usepackage{graphicx}
\usepackage{amsmath}
\usepackage{amssymb}
\usepackage{framed,multirow}
\usepackage{subfigure}
\usepackage{tabu}
\usepackage{bbding}
\usepackage{pifont}

\usepackage[breaklinks=true,bookmarks=false]{hyperref}

\cvprfinalcopy % *** Uncomment this line for the final submission

\def\cvprPaperID{****} % *** Enter the CVPR Paper ID here

\ifcvprfinal\fi
\begin{document}

%%%%%%%%% TITLE
\title{Unsupervised Instance Segmentation in Microscopy Images via Panoptic Domain Adaptation and Task Re-weighting}
% \footnote{Corresponding author}
\author{Dongnan Liu$^{1}$ \and  Donghao Zhang$^{1}$ \and Yang Song$^{2}$ \and Fan Zhang$^{3}$ \and Lauren O'Donnell$^{3}$ \and Heng Huang$^{4}$ \and Mei Chen$^{5}$ \and Weidong Cai$^{1}$ \and \\
$^{1}$School of Computer Science, University of Sydney, Australia\\$^{2}$School of Computer Science and Engineering, University of New South Wales, Australia \\ $^{3}$Brigham and Women's Hospital, Harvard Medical School, USA \\$^{4}$Department of Electrical and Computer Engineering, University of Pittsburgh, USA\\$^{5}$Microsoft Corporation, USA \\
{\tt\small \{dliu5812, dzha9516\}@uni.sydney.edu.au, yang.song1@unsw.edu.au} \\ 
{\tt\small \{fzhang, odonnell\}@bwh.harvard.edu, henghuanghh@gmail.com}\\
{\tt\small may4mc@gmail.com, tom.cai@sydney.edu.au}
}
% {\tt\small dliu5812@uni.sydney.edu.au}
% {\tt\small \{dliu5812, dzha9516\}@uni.sydney.edu.au, yang.song1@unsw.edu.au, \{fzhang, odonnell\}@bwh.harvard.edu} \\ 
% {\tt\small henghuanghh@gmail.com, may4mc@gmail.com, tom.cai@sydney.edu.au}

\maketitle
%\thispagestyle{empty}

%%%%%%%%% ABSTRACT
\begin{abstract}
Unsupervised domain adaptation (UDA) for nuclei instance segmentation is important for digital pathology, as it alleviates the burden of labor-intensive annotation and domain shift across datasets. In this work, we propose a Cycle Consistency Panoptic Domain Adaptive Mask R-CNN (CyC-PDAM) architecture for unsupervised nuclei segmentation in histopathology images, by learning from fluorescence microscopy images. More specifically, we first propose a nuclei inpainting mechanism to remove the auxiliary generated objects in the synthesized images. Secondly, a semantic branch with a domain discriminator is designed to achieve panoptic-level domain adaptation. Thirdly, in order to avoid the influence of the source-biased features, we propose a task re-weighting mechanism to dynamically add trade-off weights for the task-specific loss functions. Experimental results on three datasets indicate that our proposed method outperforms state-of-the-art UDA methods significantly, and demonstrates a similar performance as fully supervised methods. 
\end{abstract}

%%%%%%%%% BODY TEXT
\section{Introduction}

\begin{figure}[t]
\centering
\includegraphics[width=0.46\textwidth]{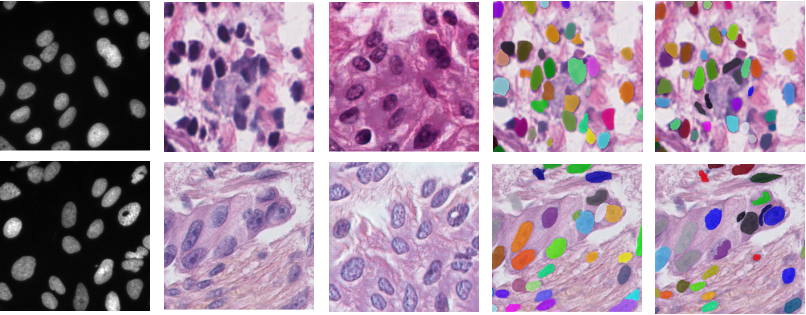} % Reduce the figure size so that it is slightly narrower than the column.
\resizebox{0.99\linewidth}{!}{%
        \begin{tabu} {  X[c]  X[c]  X[c]  X[c]  X[c]}
        (a) & (b) & (c) & (d) & (e)
        \end{tabu}}
\caption{Example images of our proposed framework. (a) fluorescence microscopy images; (b) real histopathology images; (c) our synthesized histopathology images; (d) nuclei segmentation generated by our proposed UDA method; (e) ground truth.}
\label{intro}
\end{figure}

Nuclei instance segmentation in histopathology images is an important step in the digital pathology workflow. Pathologists are able to diagnose and prognose cancers according to mitosis counts, the morphological structure of each nucleus, and spatial distribution of a group of nuclei \cite{elston1991pathological,le1989prognostic,clayton1991pathologic,basavanhally2011multi,nawaz2016computational}. Currently, supervised learning-based methods for nuclei instance segmentation are prevalent as they are efficient while preserving high accuracy \cite{kumar2017dataset,naylor2018segmentation,chen2017dcan,graham2019hover,mahmood2018deep,zhang2018panoptic,liu2019nuclei,liu2020cell}. However, their performance heavily relies on large-scale training data, which requires expertise for annotation. This process is time-consuming and labor-intensive due to the complicated cellular structures, as shown in Fig.~\ref{intro}(b), and large image sizes. For example, annotating a histopathology dataset with $50$ images and $12$M pixels costs a pathologist $120$ to $230$ hours \cite{hou2019robust}. Moreover, in real clinical studies, even one whole slide image in $40\times$ objective magnification contains $1$B pixels \cite{gutman2013cancer}. Therefore, investigating methods without depending on histopathology annotations is necessary. It can help pathologists to reduce the workload, and tackle the issue of lacking histopathology annotations.

The recently proposed unsupervised domain adaptation (UDA) methods tackle this issue by conducting supervised learning on the source domain and obtain a good performance model for the target domain without annotations \cite{pan2009survey,ganin2014unsupervised,tzeng2017adversarial}. Currently, UDA reduces distances between the distribution of feature maps of the source and target domains. In addition, some other methods focus on the pixel-to-pixel translation from the source domain images to the target ones, for aligning cross-domain image appearances \cite{isola2017image,zhu2017unpaired}. For these methods, there still remain some differences in the distributions between the synthesized and real images, due to the imperfect translations \cite{hoffman2017cycada,chen2019synergistic,kim2019diversify}.
 
% This can be done by adversarial learning on the intermediate feature maps \cite{ganin2014unsupervised,tzeng2017adversarial}, or the final model output \cite{vu2019advent,chen2018domain}. Among these methods, the feature extractors in the models are forced to generate domain-invariant features. 

% directly treating these images as the target domain images to train the task-specific CNN models causes new domain bias issues. A

To incorporate the benefits of the image translation and the UDA methods, several works have been proposed to learn the domain-invariant features between the target and the synthesized target-like images \cite{hoffman2017cycada,kim2019diversify,chen2019synergistic}. Such methods achieve state-of-the-art performance on UDA classification, object detection, and semantic segmentation tasks. However, currently there is a lack of UDA methods specifically designed for instance segmentation, and directly extending the existing UDA methods on object detection \cite{chen2018domain,kim2019diversify,he2019multi} to the UDA nuclei instance segmentation task still suffers from challenges. First, existing UDA object detection methods focus on alleviating the domain bias at the image level (image contrast, brightness, etc.) and the instance level (object scale, style, etc.) \cite{kim2019diversify,chen2018domain,he2019multi}. They ignore the domain shift at the semantic level, such as the relationship between the foreground and background, and the spatial distribution of the objects. Second, these UDA object detection methods are multi-task learning paradigms, which optimize different loss functions simultaneously. If the feature extractors fail to generate domain-invariant features in some training iterations, then back-propagating the weights according to the task loss functions in these iterations causes the model bias towards the source domain.

% To this end, employing a task-specific UDA model on the real and the synthesized histopathology images are effective for unsupervised nuclei segmentation. Currently, detection based instance segmentation methods are proposed to process every single object in the corresponding region of interest (ROI) \cite{he2017mask,liu2018path,huang2019mask,chen2019hybrid}.

To solve the aforementioned problems in UDA nuclei instance segmentation tasks in histopathology images, we propose a Cycle-Consistent Panoptic Domain Adaptive Mask R-CNN (CyC-PDAM) model. As none of the previous UDA methods are specially designed for instance segmentation, we extend the CyCADA \cite{hoffman2017cycada} to an instance segmentation version based on Mask R-CNN \cite{he2017mask}, as our baseline. In our CyC-PDAM, we firstly propose a simple nuclei inpainting mechanism to remove the auxiliary nuclei in the synthesized histopathology images. Second, inspired by the panoptic segmentation architectures \cite{kirillov2019panoptic,kirillov2019panopticfpn}, we propose a semantic-level adaptation module for domain-invariant features based on the relationship between the foreground and the background. By reconciling the domain-invariant features at the semantic and instance levels, our proposed CyC-PDAM achieves panoptic-level domain adaptation. Furthermore, a task re-weighting mechanism is proposed to reset the importance for each task loss. During training, the specific task losses are down-weighted if the features for task predictions are not domain-invariant and source-biased, and up-weighted if the features are hard to differentiate. 

% , with the image-level and instance-level adaptation (detailed illustrations in the following sections)
%  which unify the semantic and instance segmentation,
% according to the predictions of each domain classifier.
% , by keeping the original CycleGAN and changing the original adversarial domain adaptation model to a domain adaptive Mask-CNN

To prove the effectiveness of our proposed CyC-PDAM architecture, experiments have been conducted on three public datasets for unsupervised nuclei instance segmentation of histopathology images on two different datasets by unsupervised domain adaptation from a fluorescence microscopy image dataset. Unlike histopathology images, no structures are similar to the nuclei in the background of fluorescence microscopy images, due to the differences between image acquisition techniques, as shown in Fig.~\ref{intro}(a). It is much easier to obtain manual annotation for the fluorescence microscopy images compared with histopathology images, therefore it is chosen as our source domain.

Our contribution is summarized as follows: (1) We propose a CyC-PDAM model for UDA nuclei instance segmentation in histopathology images. To our best knowledge, this is the first UDA instance segmentation method. (2) A simple nuclei inpainting mechanism is proposed to remove false-positive objects in the synthesized images. (3) Our CyC-PDAM produces domain-invariant features at the panoptic level, by integrating the instance-level adaptation with a newly proposed semantic-level adaptation module. (4) A task re-weighting mechanism is proposed to alleviate the domain bias towards the source domain. (5) Compared with state-of-the-art UDA methods, our proposed CyC-PDAM paradigm outperforms them by a large margin. Moreover, it achieves competitive performance compared with state-of-the-art fully supervised methods for nuclei segmentation.

% \begin{itemize}
% \item We propose a CyC-PDAM model for UDA nuclei instance segmentation in histopathology images. To our best knowledge, this is the first UDA instance segmentation method for medical image analysis.
% \item A simple nuclei inpainting mechanism is proposed to remove false-positive objects in the synthesized images.
% \item  Our CyC-PDAM produces domain-invariant features in panoptic-level, by integrating the instance-level adaptation with a newly proposed semantic-level adaptation module.
% \item A task re-weighting mechanism is proposed to alleviate the domain bias to the source domain when the features are hard to confuse the domain classifiers.
% \item Compared with several state-of-the-art UDA methods, our proposed CyC-PDAM paradigm outperforms them by a large margin. Moreover, it achieves competitive performances compared with state-of-the-art fully supervised methods for nuclei segmentation.
% \end{itemize}

\section{Related Work}
\begin{figure*}[t]
\centering
\includegraphics[width=0.85\textwidth]{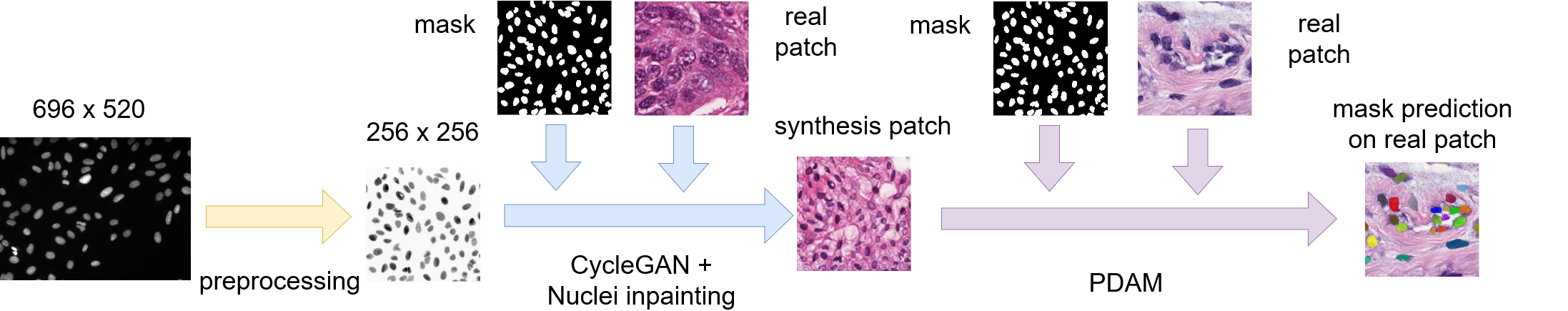} % Reduce the figure size so that it is slightly narrower than the column.
\caption{Overall architecture for our proposed CyC-PDAM architecture. The annotations of the real histopathology patches are not used during training.}
\label{overall}
\end{figure*}

\subsection{Domain Adaptation for Natural Images}
Domain adaptation aims at transferring the knowledge learned from one labeled domain to another without annotation \cite{pan2009survey}. Recently, UDA methods have reduced the cross-domain discrepancies based on the content in the feature level and the appearance in the pixel level. For the feature-level adaptation, adversarial learning for domain-invariant features \cite{ganin2014unsupervised,tzeng2017adversarial}, Maximum Mean Discrepancy minimization (MMD) \cite{long2015learning}, local pattern alignment \cite{wen2019exploiting}, and cross-domain covariance alignment \cite{sun2016return} are widely employed for classification tasks. In addition, domain adaptation is further employed for other tasks such as semantic segmentation~\cite{vu2019advent,li2019bidirectional} and object detection~\cite{chen2018domain,kim2019diversify,inoue2018cross,wang2019few}. In semantic segmentation tasks, the segmentation results are forced to be domain-invariant, together with intermediate feature maps \cite{li2019bidirectional,vu2019advent,tsai2018learning}. Additionally, ADVENT \cite{vu2019advent} further minimized the Shannon entropy for the semantic segmentation predictions in source and target domains to alleviating the cross-domain discrepancy. For object detection, a domain adaptive Faster R-CNN \cite{ren2015faster}, consisting of the image- and instance-level adaptions, was usually proposed for domain-invariant features of the whole image and each object~\cite{chen2018domain,kim2019diversify,he2019multi}. On the other hand, image-to-image translation addresses the domain adaptation problems in the pixel level by generating target-like images and training task-specific fully supervised models on them \cite{liu2017unsupervised,huang2018multimodal,isola2017image,zhu2017unpaired,mahmood2018deep,park2019semantic}. However, domain bias still exists because of imperfect translation. Moreover, several methods have been proposed to align the feature-level adaptation with the pixel-level one, by learning domain-invariant features between the target images and the synthesized images \cite{hoffman2017cycada,kim2019diversify,chen2019synergistic}.  
\subsection{Domain Adaptation for Medical Images}

% \begin{figure}[t]
% \centering
% \includegraphics[width=0.48\textwidth]{dsig-9} % Reduce the figure size so that it is slightly narrower than the column.
% \caption{Overall architecture for Dual-Stage Image Generator (DSIG). 
% % $G_{cyc1}$ and $G_{cyc2}$ are CycleGAN with the same architecture as ~\cite{zhu2017unpaired}.
% }
% \label{datagen}
% \end{figure}

Unsupervised domain adaptation for medical image analysis has rarely been explored \cite{ren2018adversarial,zhang2018task,chen2019synergistic,huang2017epithelium,hou2019robust}. \cite{ren2018adversarial} and \cite{huang2017epithelium} solve the UDA histopathology images classification problems with GAN based architectures. In addition, DAM \cite{dou2018unsupervised} is proposed to generate domain-invariant intermediate features and model predictions, for UDA semantic segmentation in CT images. With the help of cycle-consistency reconstruction, TD-GAN \cite{zhang2018task} and SIFA~\cite{chen2019synergistic} are proposed for semantic segmentation on different medical images, with both pixel- and feature-level adaptations. However, none of them is designed for UDA nuclei instance segmentation. Even though Hou \etal~\cite{hou2019robust} proposed to train a GAN based refiner and a nuclei segmentation model with the synthesized histopathology images for unsupervised nuclei instance segmentation, their paradigm only contains pixel-level adaptation and is still not capable for minimizing the domain gap in the feature level. In this work, we therefore propose a CyC-PDAM paradigm for UDA nuclei instance segmentation, which alleviates the domain bias issue in the pixel and feature levels.

\section{Methods}

Our proposed architecture is based on CyCADA and we fuse CyCADA with the instance segmentation framework Mask R-CNN. Furthermore, we improve it with nuclei inpainting mechanism, panoptic-level domain adaptation, and task re-weighting mechanism. Fig.~\ref{overall} illustrates the overall architecture of our approach. 

\subsection{CyCADA with Mask R-CNN \label{baseline-sec}}

\begin{table}[ht]
\centering
\resizebox{0.8\linewidth}{!}{%
\begin{tabular}{|l|l|l|}
\hline
Name  & Hyperparamaters & Output size  \\ 
\hline
Input            &  & $256 \times 8 \times 8$   \\ \hline
Conv1    & $k = (3, 3), s = 1, p = 1$  & $256 \times 8 \times 8$  \\ \hline
Conv2    & $k = (3, 3), s = 1, p = 1$  & $512 \times 8 \times 8$  \\ \hline
Conv3    & $k = (3, 3), s = 1, p = 1$  & $512 \times 8 \times 8$  \\ \hline
Conv4    & $k = (1, 1), s = 1, p = 0$  & $2 \times 8 \times 8$  \\ \hline
\end{tabular}}
\caption{The parameters for each block in the image-level discriminator for PDAM. $k$, $s$, and $p$ denote the kernel size, stride, and padding of the convolution operation, respectively.}
\label{dimg}
\end{table}

\begin{figure*}[t]
\centering
\includegraphics[width=0.80\textwidth, height=0.3\textwidth]{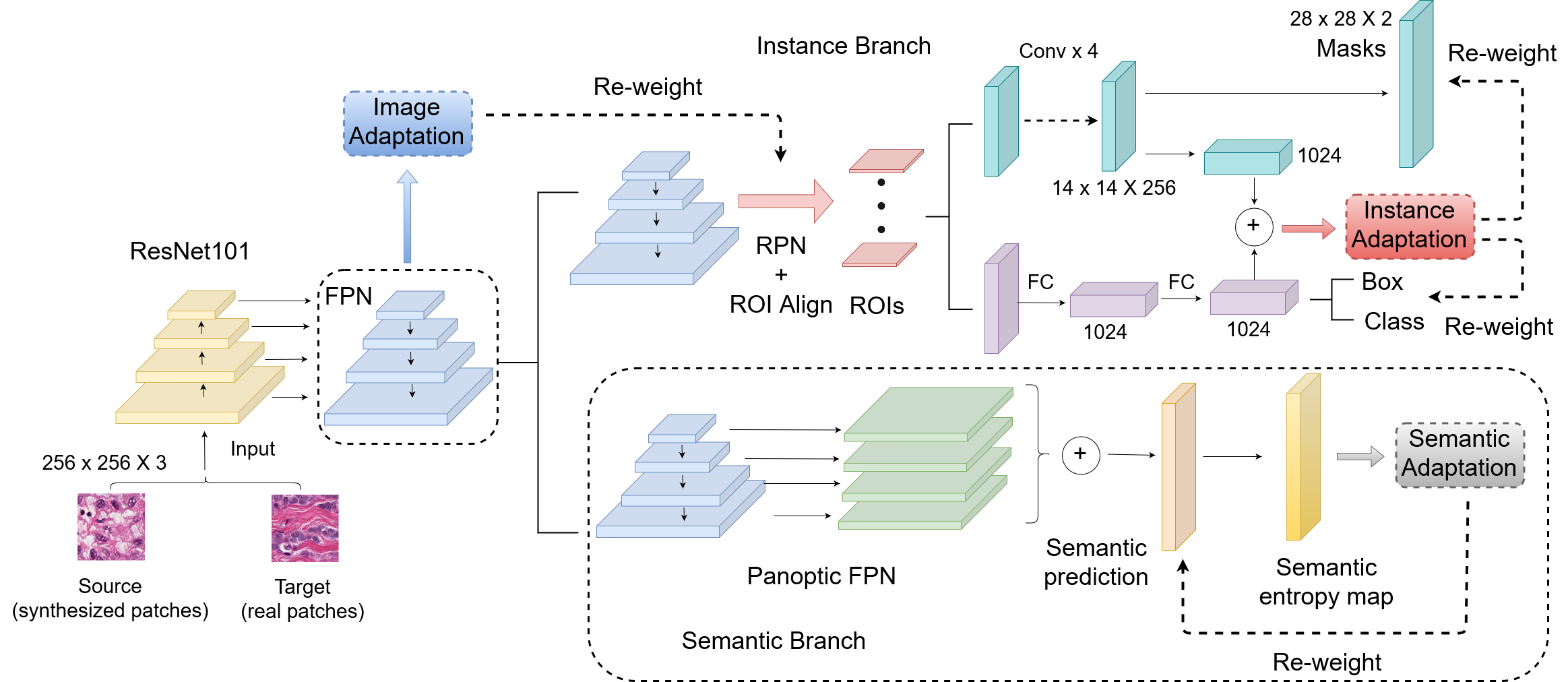} % Reduce the figure size so that it is slightly narrower than the column.
\includegraphics[width=0.80\textwidth, height=0.1\textwidth]{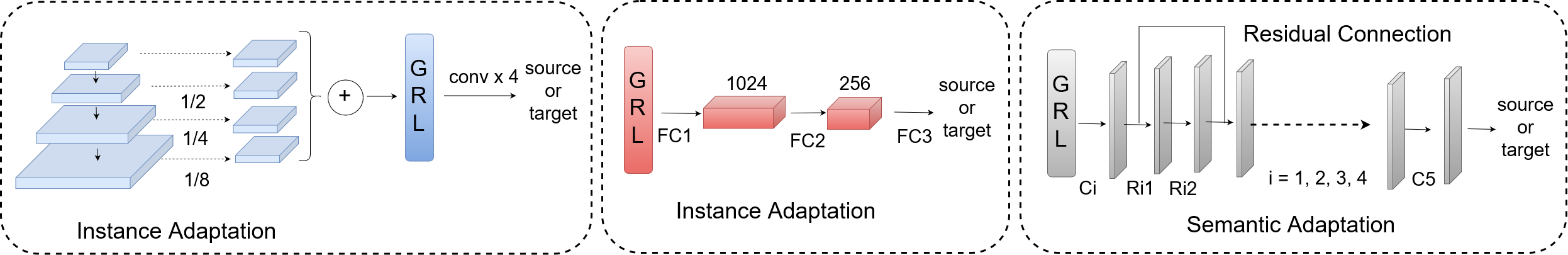}
\caption{Detailed illustration of Panoptic Domain Adaptive Mask R-CNN (PDAM). $Ci$ and $FC$ represent a convolution layer, and a fully connected layer, respectively.  $Ri1$ and $Ri2$ mean the first and second convolutional layers in the $ith$ residual block, respectively. $ReLU$ and normalization layers after each convolutional block are omitted for brevity. }
\label{pda}
\end{figure*}

As there is no UDA architectures targeting instance-level segmentation, we firstly design a domain adaptive Mask R-CNN. The backbone of the Mask R-CNN in this work is constructed with ResNet101 \cite{he2016deep} and Feature Pyramid Network (FPN) \cite{lin2017feature}. Inspired by the previous UDA methods for object detection \cite{chen2018domain,kim2019diversify}, we add one discriminator after FPN for the image-level adaptation, and the other after the instance branch for instance-level adaptation, as shown in Fig.~\ref{pda}. For the image-level adaptation, the multi-resolution feature maps of the FPN output are firstly down-sampled to the size $8 \times 8$ with average pooling, and then summed together for the image-level discriminator. The image-level discriminator consists of $4$ convolutional layers (details in Table~\ref{dimg}) and a gradient reversal layer (GRL) for adversarial learning. In the instance-level adaptation, the $14 \times 14 \times 256$ feature map in the mask branch is down-scaled to the size $2 \times 2 \times 256$ with average pooling and then resized to $1024 \times 1$, to sum with the $1024 \times 1$ feature from the bounding box branch. The instance-level discriminator consists of $3$ fully connected layers and a GRL, whose input is the summation of features mentioned above.

\subsection{Nuclei Inpainting Mechanism}

\begin{figure}[t]
\centering
\includegraphics[width=0.3\textwidth]{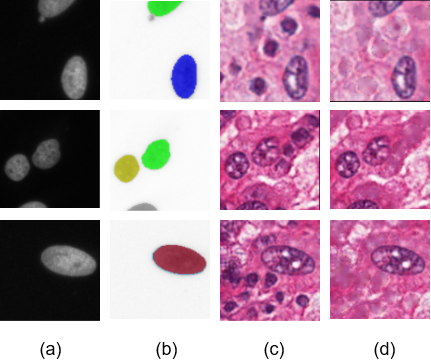} % Reduce the figure size so that it is slightly narrower than the column.
\caption{Visual results for the effectiveness of nuclei inpainting mechanism. (a) original fluorescence microscopy patches; (b) corresponding nuclei annotations; (c) initial synthesized images from CycleGAN; (d) final synthesized images after nuclei inpainting mechanism.}
\label{dsig-vis}
\end{figure}

Even though CycleGAN is effective for synthesizing histopathology-like images, due to the large domain gap and nuclei number incompatibility between the source and target domains, the label space for the generated images sometimes changes after transferring from the source domain. For example, there are redundant and undesired nuclei in the synthesized images shown in Fig.~\ref{dsig-vis}. If these images are directly used to train the task-specific CNN with the original labels, the model is forced to regard redundant nuclei as background, even though they appear as real nuclei.

Therefore, we propose an auxiliary nuclei inpainting mechanism to remove the nuclei which only appear in the synthesized images without corresponding annotations.
Denoting a raw synthesized histopathology image by CycleGAN as $S_{raw}$ and its corresponding mask as $M$, we first obtain the mask predictions $M_{aux}$ of all the auxiliary generated nuclei, formulated as:

\begin{equation}
\begin{aligned}
M_{aux} = (otsu(S_{raw}) \cup M) - M
%\\ F_{i-1}[x_{i}: x_{i} + w_{i}, y_{i}: y_{i} + h_{i}])
\end{aligned}
\end{equation}
where $ostu(S_{raw})$ represents a binary segmentation method for $S_{raw}$ based on Otsu threshold. In $M_{aux}$, only auxiliary nuclei without annotation is labeled. Then, we get the newly synthesized image $S_{inp}$ after removing these nuclei, which can be represented as:

\begin{equation}
\begin{aligned}
S_{inp} =inp(S_{raw}, M_{aux}) 
%\\ F_{i-1}[x_{i}: x_{i} + w_{i}, y_{i}: y_{i} + h_{i}])
\end{aligned}
\end{equation}
where $inp$ is a fast marching based method for inpainting objects \cite{telea2004image}, by replacing the pixel values for the auxiliary nuclei labeled in $M_{aux}$ with them for the unlabeled background. Fig.~\ref{dsig-vis} illustrates the visual effectiveness of our proposed nuclei inpainting mechanism. However, some background materials are labeled as false positive predictions in $M_{aux}$. Directly inpainting them makes the texture and appearance of synthesized images unrealistic, and enlarges the domain gap between the synthesized and real images. However, the image-level adaptation is able to address this issue by alleviating the domain bias on global visual information, such as curve, texture, and illumination. Our nuclei inpainting mechanism is time-efficient, which takes $0.09$ second to process one single $256 \times 256$ synthesized histopathology patch, on average.

\subsection{Panoptic Level Domain Adaptation}

We define the semantic-level features of an image as the relationship between its foreground and background. In addition to the image- and feature-level domain bias, the domain shift at the semantic level also exists. Due to the differences in the nuclei objects and background between the synthesized and real histopathology images, domain adaptive Mask R-CNN mentioned in Sec.~\ref{baseline-sec} suffers from domain bias in the semantic-level features, as the Mask R-CNN only focuses on the local features for each object and lacks a semantic view of the whole image. Inspired by the previous panoptic segmentation architecture, which unified the semantic and instance segmentation to process the global and local features of the images, we propose a semantic-level adaptation to induce the model to learn domain-invariant features based on the relationship between the foreground and background. By incorporating the semantic- and instance-level adaptation, our panoptic domain adaptive method reduces the cross-domain discrepancies in a global and local view.

As shown in Fig.~\ref{pda}, a semantic branch for semantic segmentation prediction is added to the output of the FPN. Our semantic branch has the same implementation as \cite{kirillov2019panopticfpn}. As the fluorescence microscopy images and histopathology images can both be acquired from tissue samples and they can show complementary and correlated information, the semantic segmentation label spaces of the synthesized and real histopathology images have a strong similarity. In addition, aligning the cross-domain entropy distributions helps to minimize the entropy prediction in the target domain, which makes the model suitable for the target images \cite{vu2019advent}. Therefore, we use the Shannon entropy \cite{shannon1948mathematical} of the softmax semantic predictions to induce the domain-invariant features to learn at the semantic level. Denoting the softmax semantic prediction as $P$ and $P \in (0,1)$, its Shannon entropy is defined as: $-plog(p)$.

Fig.~\ref{pda} and Table~\ref{dsem} indicate the detailed structure of the discriminator for semantic level adaptation. We employ residual connected CNN blocks to avoid gradient vanishing \cite{he2016deep,he2016identity}. To make the adversarial learning more stable, instead of bilinear interpolation, we use stride convolutional layers for upsampling. Finally, the domain label is predicted as a $16 \times 16$ patch. Due to the small mini-batch size, the patch-based domain label prediction increases the number of training samples, to avoid overfitting.

\begin{table}[ht]
\centering
\resizebox{0.8\linewidth}{!}{%
\begin{tabular}{|l|l|l|}
\hline
Name  & Hyperparamaters & Output size  \\ 
\hline
Input            &  & $2 \times 256 \times 256$   \\ \hline
C1    & $k = (7, 7), s = 2, p = 3$  & $64 \times 128 \times 128$  \\ \hline
R11 and R12           & $k = (3,3), s = 1, p = 1$ & $64 \times 128 \times 128$    \\ \hline
C2    & $k = (5, 5), s = 2, p = 2$  & $128 \times 64 \times 64$  \\ \hline
R21 and R22           & $k = (3,3), s = 1, p = 1$ & $128 \times 64 \times 64$    \\ \hline
C3    & $k = (5, 5), s = 2, p = 2$  & $256 \times 32 \times 32$  \\ \hline
R31 and R32           & $k = (3,3), s = 1, p = 1$ &$256 \times 32 \times 32$  \\ \hline
C4    & $k = (5, 5), s = 2, p = 2$  & $512 \times 16 \times 16$   \\ \hline
R41 and R42           & $k = (3,3), s = 1, p = 1$ &$512 \times 16 \times 16$  \\ \hline
C5   & $k = (1,1), s = 1, p = 0$ & $2 \times 16 \times 16$   \\ \hline
  Output   &  & $2 \times 16 \times 16$   \\ \hline
\end{tabular}}
\caption{The parameters for each block in the semantic-level discriminator for PDAM. $k$, $s$, and $p$ follow the same convention as in Table \ref{dimg}.}
\label{dsem}
\end{table}

\subsection{Task Re-weighting Mechanism}
 
In the previous UDA methods, the task-specific loss functions (segmentation, classification, and detection) are based on the source domain predictions. Even though several adversarial domain discriminators are employed to ensure the predicted feature maps are domain-invariant, the cross-domain discrepancies of these feature maps are still large in some training iterations, where the features are far from the decision boundaries of the domain discriminators. If the task-specific losses are updated to optimize the models with these easily-distinguished features, the models will bias towards the source images when testing it with the target data. To this end, we propose a task re-weighting mechanism to add a trade-off weight for each task-specific loss function according to the prediction of the domain discriminator. Denote the probability of the feature map before the final task prediction belonging to the source and target domains as $p_{s}$ and $p_{t}$, respectively, and the task-specific loss function as $L$, then the re-weighted task-specific loss $L_{rw}$ is:

\begin{equation}
\begin{aligned}
L_{rw} = min (\frac{p_{t}}{p_{s}} , \beta)L = min(\frac{1-p_{s}}{p_{s}}, \beta) L
\end{aligned}
\label{taskrw}
\end{equation}
where $\beta$ is a threshold value to avoid the $\frac{1-p_{s}}{p_{s}}$ becoming large and making the model collapse, when $p_{s} \to 0$. According to Eq.~\ref{taskrw}, if the feature map deciding the task prediction belongs to the source domain ($p_{s} \to 1$), the loss function is then down-weighted, to alleviate the source-bias feature learning of the model. As illustrated in Fig.~\ref{pda}, the loss function for the region proposal network (RPN), semantic branch, and the instance branch are re-weighted by the prediction at the image-, semantic-, and instance-level domain discriminators, respectively.

\subsection{Network Overview and Training Details}

In our proposed CyC-PDAM, the CycleGAN has the same implementation as its original work \cite{zhu2017unpaired}. When training the CycleGAN, the initial learning rate was set to $0.0001$ for the first $1/2$ of the total training iterations, and linearly decayed to $0$ for the other $1/2$. 

The PDAM is trained with a batch size of $1$ and each batch contains $2$ images, one from the source and the other from the target domain. Due to the small batch size, we replace traditional batch normalization layers with group normalization \cite{wu2018group} layers, with the default group number $32$ as in \cite{wu2018group}. 

The overall loss function of PDAM is defined as:

\begin{equation}
\begin{aligned}
    L_{pdam} & = \alpha_{img} L_{rpn}  + \alpha_{ins} L_{det} + \alpha_{sem} L_{(sem-seg)}  \\
    & + \alpha_{da}(L_{(img-da)} + L_{(sem-da)} + L_{(ins-da)}) 
\end{aligned}
\end{equation}
where $L_{rpn}$ is the loss function for the RPN, $L_{det}$ is the loss of class, bounding box, and instance mask prediction of Mask R-CNN, $L_{(sem-seg)}$ is the cross entropy loss for semantic segmentation, $L_{(img-da)}$, $L_{(sem-da)}$ and $L_{(ins-da)}$ are cross entropy losses for domain classification at image, semantic and instance levels. $\alpha_{img}$, $\alpha_{ins}$, and $\alpha_{isem}$ are calculated according to Eq.~\ref{taskrw} for task re-weighting. In our experiment, we set $\beta$ as $2$. $\alpha_{da}$ is updated as:

\begin{equation}
\begin{aligned}
    \alpha_{2} = \frac{2}{1 + exp(-10t)} - 1
\end{aligned}
\end{equation}
where $t$ is the training progress and $t \in [0, 1]$. Thus $\alpha_{da}$ is gradually changed from $0$ to $1$, to avoid the noise from the unstable domain discriminators in the early training stage.

During training, the PDAM is optimized by SGD, with a weight decay of $0.001$ and a momentum of $0.9$. The initial learning rate is $0.002$, with linear warming up in the first $500$ iterations. The learning rate is then decreased to $0.0002$ when it reaches $3/4$ of the total training iteration. During inference, only the original Mask R-CNN architecture is used with the adapted weight and all of the hyperparameters for testing are fine-tuned on the validation set. All of our experiments were implemented with Pytorch \cite{paszke2017automatic}, on two NVIDIA GeForce 1080Ti GPUs. 

\section{Experiments}

\begin{table*}[!t]
\centering
%\begin{tabular}{|p{2.25cm}|p{2cm}|l|p{4cm}|p{3cm}|p{2cm}|}
\resizebox{0.85\linewidth}{!}{%
\begin{tabular}{|c|c|c|c|c|c|c|}
 \hline
  &    \multicolumn{3}{c|}{$BBBC039 \to Kumar$}  &  \multicolumn{3}{c|}{$BBBC039 \to TNBC$}  \\
   \hline
 Methods  &    AJI &  Pixel-F1 & Object-F1 &    AJI &  Pixel-F1 & Object-F1\\
  \hline
 CyCADA \cite{hoffman2017cycada} &    $0.4447 \pm 0.1069 $ &  $0.7220 \pm 0.0802 $ & $0.6567 \pm 0.0837 $ &   $0.4721 \pm 0.0906$ &  $0.7048 \pm 0.0946 $ & $0.6866 \pm 0.0637$\\
  \hline
Chen \etal \cite{chen2018domain}  &    $0.3756 \pm 0.0977 $ &  $0.6337 \pm 0.0897 $ & $0.5737 \pm 0.0983$ &    $0.4407 \pm 0.0623 $ &  $0.6405 \pm 0.0660$ & $0.6289 \pm 0.0609 $\\
 \hline
SIFA \cite{chen2019synergistic} &    $0.3924 \pm 0.1062 $ &  $0.6880 \pm 0.0882$ & $0.6008 \pm 0.1006$ &    $0.4662 \pm 0.0902 $ &  $0.6994 \pm 0.0942$ & $0.6698 \pm 0.0771$\\
 \hline
 DDMRL \cite{kim2019diversify} &    $0.4860 \pm 0.0846 $ &  $0.7109 \pm 0.0744$ & $0.6833 \pm 0.0724$ &    $0.4642 \pm 0.0503 $ &  $0.7000 \pm 0.0431$ & $0.6872 \pm 0.0347$\\
 \hline
Hou \etal \cite{hou2019robust} &    $0.4980 \pm 0.1236$ &  $0.7500 \pm 0.0849 $ & $0.6890 \pm 0.0990 $ &    $0.4775 \pm 0.1219 $ &  $0.7029 \pm 0.1262 $ & $0.6779 \pm 0.0821 $\\
 \hline
Proposed  &    $ \textbf{0.5610} \pm \textbf{0.0718} $ & $  \textbf{0.7882} \pm \textbf{0.0533}  $ & $ \textbf{0.7483} \pm \textbf{0.0525} $ &   $ \textbf{0.5672} \pm \textbf{0.0646} $ &  $ \textbf{0.7593} \pm \textbf{0.0566}$ & $ \textbf{0.7478} \pm \textbf{0.0417}  $\\
 \hline
\end{tabular}}
\caption{In comparison with other unsupervised methods on both two histopathology datasets.}
\label{cmp-exp}
\end{table*}

\subsection{Datasets Description and Evaluation Metrics}

Our proposed architecture was validated on three public datasets, referred to as Kumar \cite{kumar2017dataset}, TNBC \cite{naylor2018segmentation}, and BBBC039V1 \cite{ljosa2012annotated}, respectively. Among them, Kumar and TNBC are histopathology datasets, while BBBC039V1 is a fluorescence microscopy dataset. Kumar was acquired from The Cancer Genome Atlas (TCGA) at $40 \times$ magnification, containing $30$ annotated $1000 \times 1000$ patches from $30$ whole slide images of different patients. All these images are from $18$ different hospitals and $7$ different organs (breast, liver, kidney, prostate, bladder, colon, and stomach). In contrast to the disease variability in Kumar, the TNBC dataset especially focuses on Triple-Negative Breast Cancer (TNBC) \cite{naylor2018segmentation}. In TNBC, there are $50$ annotated $512 \times 512$ patches from $11$ different patients from the Curie Institute at $40 \times$ magnification. BBBC039V1 is about U2OS cells under a high-throughput chemical screen \cite{ljosa2012annotated}. It contains $200$ $520 \times 696$ images about bioactive compounds, with the DNA channel staining of a single field of view.

For evaluation, we employ three commonly used pixel- and object-level metrics. Aggregated Jaccard Index (AJI) is an extended Jaccard Index for object-level evaluation \cite{kumar2017dataset}, and object-level F1 score is the average harmonic mean between the precision and recall for each object. For pixel-level evaluation, we employ pixel-level F1 score for binarization predictions.

\subsection{Experiment Setting}

We conducted our experiments on two nuclei segmentation tasks: adapting from BBBC039V1 to Kumar, and from BBBC039V1 to TNBC. As the source domain in two experiments, $100$ training images and $50$ validation images from BBBC039V1 are used, following the official data split\footnote{~\url{https://data.broadinstitute.org/bbbc/BBBC039/}}. The annotations for Kumar and TNBC are not used during training the UDA architecture, only for evaluation.

The preprocessing for source fluorescence microscopy images has $3$ steps. First, all images are normalized into range $[0, 255]$. Second, $10K$ patches in size $256 \times 256$ are randomly cropped from the $100$ training images, with data augmentation including rotation, scaling, and flipping to avoid overfitting. Third, the patches with fewer than $3$ objects are removed. For better synthesizing target-like histopathology images, we finally inverse the pixel value of foreground nuclei and background for all source fluorescence microscopy patches. For validation, $50$ images in the BBBC039V1 validation set are transferred to synthesized histopathology images by CycleGAN and nuclei inpainting mechanism.

For the Kumar dataset as the target domain, we have the same data split as previous work in \cite{kumar2017dataset,naylor2018segmentation}, with $16$ images for training, and $14$ for testing. When training the model, totally $10K$ patches in size $256 \times 256$ are randomly cropped from the $16$ training histopathology images, with basic data augmentation including flipping and rotation, to avoid overfitting. As for TNBC, we use $8$ cases with $40$ images for training, and the remaining $3$ cases with $10$ images for testing. To train the model with TNBC, $10K$ $256 \times 256$ patches are randomly extracted from the training images with basic data augmentation including flipping and rotation. 

\subsection{Comparison Experiments}

\subsubsection{Comparison with Unsupervised Methods \label{sec-cmp}}

In this section, our proposed CyC-PDAM is compared with several state-of-the-art UDA methods, including CyCADA \cite{hoffman2017cycada}, Chen \etal \cite{chen2018domain}, SIFA \cite{chen2019synergistic}, and DDMRL \cite{kim2019diversify}. As the original CyCADA focuses on classification and semantic segmentation, we extend it with Mask R-CNN for UDA instance segmentation, as described in Sec.~\ref{baseline-sec}. Chen \etal \cite{chen2018domain} are originally for UDA object detection based on Faster R-CNN, by adapting the features at the image and instance levels. For UDA instance segmentation, we replace the original VGG16 based Faster R-CNN with the same Mask R-CNN in our architecture, and the original image- and instance-level adaptation in \cite{chen2018domain} with ours in Sec.~\ref{baseline-sec}. SIFA \cite{chen2019synergistic} is a UDA semantic segmentation architecture for CT and MR images, with a pixel- and feature-level adaptation. In our experiment, we add the watershed algorithm to separate the touching objects in the semantic segmentation prediction of SIFA, for a fair comparison. DDMRL \cite{kim2019diversify} learns multi-domain-invariant features from various generated domains for UDA object detection and it is extended for instance segmentation, in a similar way as CyCADA \cite{hoffman2017cycada} and Chen \etal \cite{chen2018domain}. In addition, we also compared with Hou \etal \cite{hou2019robust}, which is particularly designed for unsupervised nuclei segmentation in histopathology images. They trained a multi-task (segmentation, detection, and refinement) CNN architecture with their synthesized histopathology images from randomly generated binary nuclei masks.

\begin{figure}[t]
\centering
\includegraphics[width=0.49\textwidth]{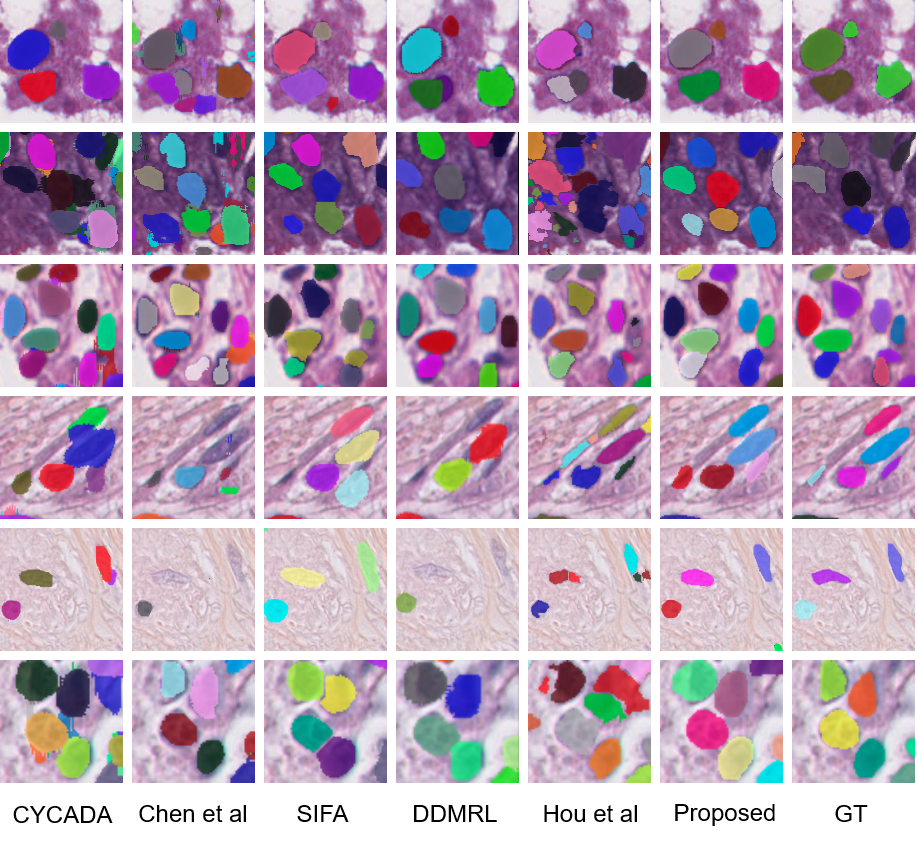} % Reduce the figure size so that it is slightly narrower than the column.
% \resizebox{0.99\linewidth}{!}{%
%         \begin{tabu} {  X[c]  X[c]  X[c]  X[c]  X[c]  X[c]} 
%         (a) & (b) & (c) & (d) & (e) & (f)
%         \end{tabu}}
\caption{Visualization result for the comparison experiments experiment. The first $3$ rows are from Kumar dataset, and the last $3$ rows are from TNBC.
% (a): CyCADA; (b): Chen et al; (c): SIFA; (d): Hou et al; (e): Propsoed; (f): Ground truth.
}
\label{cmp-vis}
\end{figure}

Table~\ref{cmp-exp} shows that our proposed method outperforms all the comparison methods by a large margin, on different histopathology datasets. In addition, the one-tailed paired t-test is employed to prove that all of our improvements are statistically significant, with all the p-values under $0.05$. Chen \etal \cite{chen2018domain} learns the domain-invariant features at the image and instance levels. However, due to the large differences between the fluorescence microscopy and real histopathology images, feature-level adaptation only is not enough to reduce the domain gap. With pixel-level adaptation on appearance, all the other methods achieve better performance. Compared with the baseline method CyCADA \cite{hoffman2017cycada}, our CyC-PDAM has a large improvement of $6 - 12 \%$, due to the effectiveness of our proposed nuclei inpainting mechanism, panoptic-level adaptation, and task re-weighting mechanism. SIFA \cite{chen2019synergistic} focuses on domain-invariant features in the image and semantic levels, with a UDA semantic segmentation structure. As there exists a large number of nuclei objects in the histopathology images, the effectiveness of SIFA is still limited without any instance-level learning or adaptation. Although DDMRL \cite{kim2019diversify} only adapts the features at the image level, its performance is still at the same level as CyCADA, by adapting knowledge across various domains. Among all the comparison methods, Hou \etal \cite{hou2019robust} achieves the second-best performance. Due to the effectiveness of panoptic-level feature adaptation and task re-weighting mechanism, our method still outperforms it under all three metrics, in both two experiments. Fig.~\ref{cmp-vis} are visualization examples of all the comparison methods.

\subsubsection{Ablation Study}

\begin{table}[!t]
\centering
%\begin{tabular}{|p{2.25cm}|p{2cm}|l|p{4cm}|p{3cm}|p{2cm}|}
\resizebox{0.85\linewidth}{!}{%
\begin{tabular}{|c|c|c|c|}
 \hline
 &    AJI & Pixel-F1 & Object-F1 \\
\hline
%  w/o adaptation & $0.3170 \pm 0.1388 $ & $0.5076 \pm 0.1781 $ & $0.4662 \pm 0.1914 $\\
% \hline

w/o NI &  $0.5042 \pm 0.1034 $ & $0.7336 \pm 0.0839$  & $0.6958 \pm 0.0832 $\\
\hline
w/o TR &  $0.4969 \pm 0.0972  $ & $0.7654 \pm 0.0678$  & $0.6923 \pm 0.0778 $ \\
\hline
w/o SEM  & $0.5046\pm 0.1065$ & $0.7470 \pm 0.0754 $ & $0.6965 \pm 0.0805  $ \\
\hline
proposed  & $0.5610 \pm 0.0718 $ & $0.7882 \pm 0.0533 $ & $0.7483 \pm 0.0525  $ \\
\hline
\end{tabular}}
\caption{Ablation study on BBBC039V1 to Kumar experiment. NI, TR, and SEM represent the nuclei inpainting mechanism, task re-weighting mechanism, and semantic branch, respectively. }
\label{tcga-abl}
\end{table}

\begin{table*}[!t]
\centering
%\begin{tabular}{|p{2.25cm}|p{2cm}|l|p{4cm}|p{3cm}|p{2cm}|}
\resizebox{0.85\linewidth}{!}{%
\begin{tabular}{|c|c|c|c|c|c|c|}
\hline
&  \multicolumn{3}{c|}{AJI}  &  \multicolumn{3}{c|}{Pixel-F1}  \\
\cline{1-7}
 Methods & seen & unseen  & all & seen & unseen  & all  \\
\cline{1-7}
\hline
CNN3 \cite{kumar2017dataset}  &$0.5154 \pm 0.0835 $ &  $0.4989 \pm 0.0806$ &  $0.5083 \pm 0.0695$ &  $0.7301 \pm 0.0590 $ &  $0.8051 \pm 0.1006 $ &  $0.7623 \pm 0.0946$ \\
\hline
DIST \cite{naylor2018segmentation} &$0.5594 \pm 0.0598$ &  $0.5604 \pm 0.0663$ &  $0.5598 \pm 0.0781$ &  $0.7756 \pm 0.0489 $ &  $0.8005 \pm 0.0538 $ &  $0.7863 \pm 0.0550$ \\
\hline
Proposed  &$0.5432 \pm 0.0477$ &  $0.5848 \pm 0.0951$ &  $0.5610 \pm 0.0982$ &  $0.7743 \pm 0.0358$ &  $0.8068 \pm 0.0698$ &  $0.7882 \pm 0.0533$ \\
\hline
Upper bound \cite{kirillov2019panopticfpn} &$0.5703 \pm 0.0480$ &  $0.5778 \pm 0.0671$ &  $0.5735 \pm 0.0855$ &  $0.7796 \pm 0.0419 $ &  $0.8007 \pm 0.0511$ &  $0.7886 \pm 0.0531$ \\
\hline
\end{tabular}}
\caption{Comparison experiments between our UDA method and fully supervised methods, for BBBC039V1 to Kumar experiment. For CNN3 and DIST, the results of object-level F1 are unknown.
}
\label{tcga-sup}
\end{table*}

\begin{figure}[t]
\centering
\includegraphics[width=0.45\textwidth]{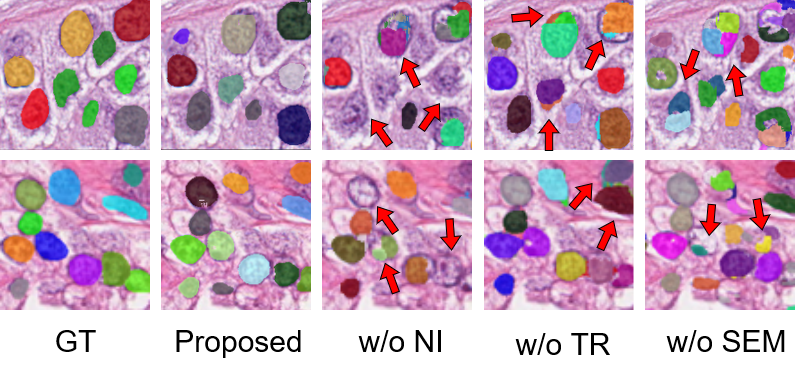} % Reduce the figure size so that it is slightly narrower than the column.
% \resizebox{1.0\linewidth}{!}{%
%         \begin{tabu} {  X[c]  X[c]  X[c]  X[c]  X[c]}
%         GT & Proposed & w/o NI & w/o TR & w/o SEM
%         \end{tabu}}
\caption{Visualization results for the ablation experiment. NI: nuclei inpainting mechanism; TR: task re-weighting mechanism; SEM: semantic branch.}
\label{abl-vis}
\end{figure}

In order to test the effectiveness of each component in our proposed CyC-PDAM, ablation experiments are conducted on the Kumar dataset. Based on our CyC-PDAM, we remove the nuclei inpainting mechanism, task re-weighting mechanism, and semantic branch for panoptic-level adaptation and train the ablated models with the same setting and dataset as Sec.~\ref{sec-cmp}. Table~\ref{tcga-abl} and Fig.~\ref{abl-vis} show the detailed results of the ablation experiment. As shown in Fig.~\ref{abl-vis}, the method without nuclei inpainting mechanism (w/o NI) tends to ignore some nuclei, which increases the false-negative predictions. Moreover, we notice that there are also false split and merged predictions for w/o NI model. It is because the increasing false negative predictions are harmful to the spatial distribution of all the objects, which further affects the effectiveness of the semantic-level adaptation. Among the predictions of the method without task re-weighting mechanism (w/o TR), there exist some objects with irregular sizes. The task re-weighting mechanism prevents the model from being influenced by the domain-specific features in the source domain, and removing it, therefore, incurs source-biased predictions. Compared with our method, the model without semantic-branch (w/o SEM) is not able to learn domain-invariant features at the semantic level, including the spatial distribution of the nuclei objects and the detailed information in the background. Therefore, there not only remain falsely split and merged predictions, but also false-positive and imperfect segmentation results. As shown in Table~\ref{tcga-abl}, the segmentation accuracy under three metrics decreases by $4 - 6\%$ after removing each module. In addition, the one-tailed paired t-test is employed to calculate the p-value between our proposed method and the other ablated methods. After adding each of the three modules, the improvements are statistically significant ($P < 0.05$), which further demonstrates the effectiveness of our proposed method.

\subsubsection{Comparison with Fully Supervised Methods}

As our data split in Kumar dataset is the same as several state-of-the-art methods for fully supervised nuclei segmentation, we compare their original reported results with ours. Table~\ref{tcga-sup} illustrates the comparison results between our proposed UDA architecture and other fully supervised methods. CNN3 \cite{kumar2017dataset} is a contour-based nuclei segmentation architecture, which considers nuclei boundaries as the third class, in addition to the foreground and background classes. DIST~\cite{naylor2018segmentation} is a regression model based on the distance map. For Panoptic FPN \cite{kirillov2019panopticfpn}, we directly train it using the same set of $16$ real histopathology patches as CNN3 and DIST and it is employed as the upper bound of our unsupervised method. The testing images for Kumar are divided into two subsets: one contains $8$ images from $4$ organs known to training set, referred to as seen, and the other contains $6$ images from $3$ organs unknown to the training set, referred to as unseen. 

As shown in Table~\ref{tcga-sup}, the performance of our proposed UDA architecture is superior to the fully supervised CNN3 and DIST. It is because our proposed method is able to process each ROI on the local level, while CNN3 and DIST only process the image at a global semantic level. By adapting the semantic-level features of the foreground and the background, the performance of our method is at the same level as the fully supervised Panoptic FPN for the pixel-level F1-score. Even though our AJI is slight lower than the fully supervised Panoptic FPN, we notice that our method works better when tested on the unseen testing set. This is because our proposed CyC-PDAM focuses on learning the domain-invariant features and avoids being influenced by the domain bias of testing images from unseen organs. These results show that, although there remains large differences between the fluorescence microscopy images and histopathology images, our proposed UDA architecture still successfully narrows the domain gap between them, and achieves even better performance compared with fully supervised methods requiring histopathology nuclei annotations.

\section{Conclusion}

In this work, we propose a CyC-PDAM architecture for UDA nuclei segmentation in histopathology images. We firstly design a baseline architecture for UDA instance segmentation, including appearance-, image-, and instance-level adaptation. Next, a nuclei inpainting mechanism is designed to remove the auxiliary objects in the synthesized images, to further avoid false-negative predictions. In the feature-level adaptation, a semantic branch is proposed to adapt the features with respect to the foreground and background, and incorporating semantic- and instance-level adaptation enables the model to learn domain-invariant features at the panoptic level. In addition, a task re-weighting mechanism is proposed to reduce the bias. Extensive experiments on three public datasets indicate our proposed method outperforms the state-of-the-art UDA methods by a large margin and reaches the same level as the fully supervised methods. From a larger perspective, the UDA instance segmentation problems are not limited to histopathology image analysis. With the promising performance close to fully supervised methods in this work, we suggest that our proposed method can also contribute to other general image analysis applications.

{\small
\bibliographystyle{ieee_fullname}
\bibliography{egbib}
}

\end{document}